\pdfoutput=1

\documentclass[11pt]{article}

\usepackage[final]{acl}

\usepackage{times}
\usepackage{latexsym}
\usepackage{enumitem}
\usepackage{float}
\usepackage{tcolorbox}
\usepackage{lipsum}
\usepackage{booktabs} 
\tcbuselibrary{breakable}

\usepackage[T1]{fontenc}

\usepackage[utf8]{inputenc}

\usepackage{microtype}
\usepackage{array}
\usepackage{amsmath}

\usepackage{inconsolata}
\usepackage{booktabs}

\usepackage{graphicx}
\usepackage{float}

\title{Beyond LLMs: A Linguistic Approach to Causal Graph Generation from Narrative Texts}

\author{
  Zehan Li \quad Ruhua Pan \quad Xinyu Pi \\
  University of California, San Diego \\
  \texttt{\{zel025, r3pan, xpi\}@ucsd.edu}
}

\begin{document}
\maketitle
\begin{abstract}
We propose a novel framework to generate causal graphs from narrative texts, bridging the gap between high-level causality and finer-grained event-specific relationships. Our approach first extracts concise, agent-centered “vertices” using an LLM-based summarization strategy. We then introduce an \textit{Expert Index}—seven linguistically grounded features—and incorporate them into a \textbf{STAC} (Situation, Task, Action, Consequence) classification model. This hybrid system (RoBERTa embeddings + Expert Index) achieves superior precision in identifying causal links compared to LLM-only baselines. Finally, we apply a structured, five-iteration prompting process to refine and construct a connected causal graph. Experiments on 100 chapters and short stories show that our method consistently outperforms GPT-4o and Claude 3.5 across key dimensions of causal graph quality, while maintaining comparable readability. The resulting open-source tool offers an interpretable and efficient solution for capturing nuanced causal chains within narrative texts.

\end{abstract}

\section{Introduction}

Causal research has historically leveraged knowledge graphs to explore relationships between events \cite{JM;_1999a}. Modern approaches, such as AI-driven causal graph generation, have gained prominence for their ability to summarize causal events at scale \cite{jaimini2022causalkg, pieper2023causalgraph}. However, current AI models largely focus on high-level causality (e.g., "HIV leads to AIDS"), and they fall short in capturing nuanced causal relationships in specific narratives, such as political events or historical occurrences\citep{Donnelly2025}. Addressing this gap, we propose a method for generating causal graphs from texts that describe discrete, event-specific narratives.

Understanding these finer-grained causal relationships is crucial for researchers and practitioners who analyze how certain events lead to tangible outcomes in areas like social movements, policy-making, and historical trends. By capturing causal links from narrative texts, stakeholders can more accurately trace the chain of events that precipitate significant changes, enabling better decision-making, deeper historical insight, and more targeted interventions. Furthermore, automated causal graph generation facilitates scalable analysis of large document collections, providing structured representations that can be easily interpreted, queried, and expanded upon.

Most existing methods for generating causal graphs follow a two-stage pipeline: (1) a Causality Finder to detect causal relations, and (2) a graph Generator to construct knowledge graphs from these relations. While effective, these methods face limitations in interpretability and accuracy, particularly when dealing with complex sentence structures or implicit causal links \citep{Kyono2024}.

Causality finders have evolved through three phases: (1) early pattern-based models that learned causal relationships from fixed sentence structures   \citep{Hidey2016} \citep{Heindorf2020} ,  (2) BERT-based approaches that addressed issues in text training but failed to account for semantic context \cite{tan2023unicausalunifiedbenchmarkrepository} \citep{Dasgupta2018} \citep{Li2020}, and (3) LLMs, which improved contextual reasoning but struggled to distinguish intricate causal relationships \cite{kiciman2024causalreasoninglargelanguage} \citep{Shen2022} \citep{Luo2024}.

In this paper, we present a novel framework that leverages linguistic feature extraction to enhance causal graph generation from narrative texts. Our approach introduces a Quaternary Classification system to categorize sentences into four components: (1) Situation, (2) Task, (3) Action, and (4) Consequences. This structured decomposition allows for more precise identification of causal links. We also propose a Neural Network model trained on these linguistic features, achieving higher accuracy and interpretability compared to LLM-based methods, with lower computational costs.

The first three dimensions of our Expert Index---Genericity, Eventivity, and Boundedness/Habituality---build on the clause-level discourse framework and annotated corpus introduced by \citet{hemmatian-etal-2021-novel} and developed in greater detail by \citet{hemmatian-borujeni-2022-taking}. In the present work, we adapt these dimensions to narrative causal-graph generation and combine them with four additional features and the STAC classification pipeline.

Our contributions are twofold: (1) We develop an open-source, end-to-end causal graph generation model that significantly improves interpretability and accuracy. (2) We introduce a Linguistics Feature system, which efficiently classifies sentences for causal graph construction, validated through experiments on various narrative texts.

\begin{figure*}[htbp]
    \centering
    \includegraphics[height=5.5cm, width=16cm]{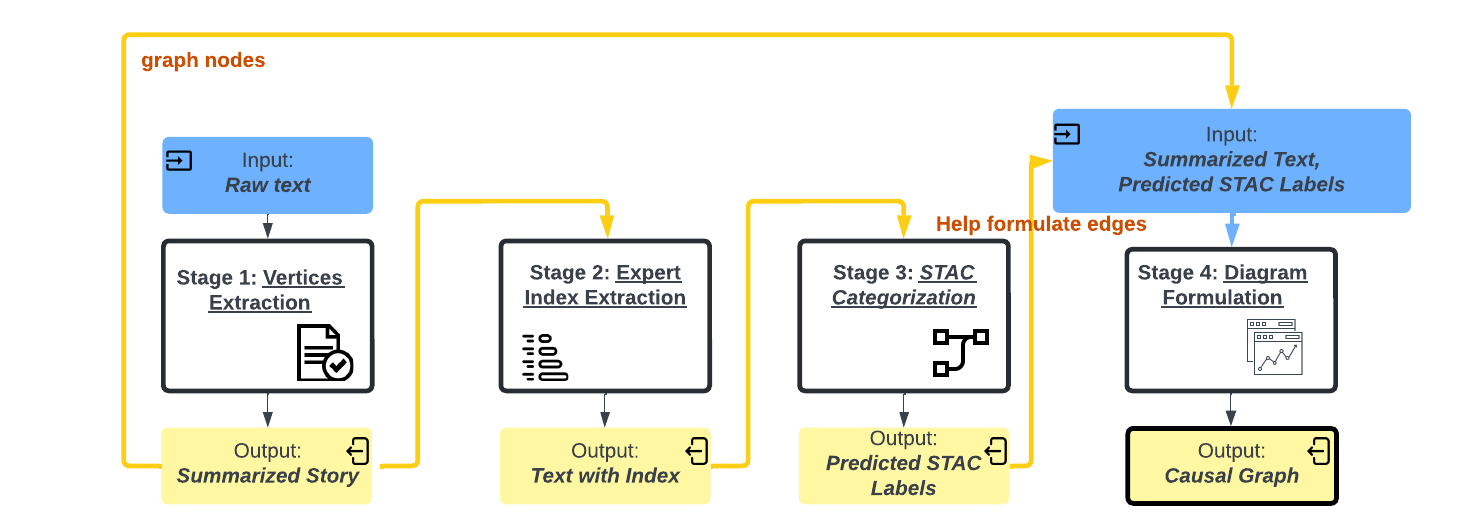}
    \captionsetup{font={small}} 
    \caption{Overview of our DA framework.  It is an end to end Model. First we input a Random Narrative Text. Then in Stage 1, we Contribute the Vertices of the Graph. And in Stage 2, we Use our Expert Index to indicate the Vertices. Next, In stage 3,we use a STAC system to label the Vertices. In STAGE 4, we use STAC Label + Vertices to complete the Causal Graph}
    \label{fig:pipeline}
\end{figure*}

\section{Problem Setting}

This paper studies the problem of causal relationship graphs as follows. Given a narrative text, such as a story by O. Henry or a piece of narrative news, we can generate its causal relationship graph containing the main causal relationships. More specifically, when we input a set of narrative sentences \( S = \{s_1, s_2, \ldots, s_n\} \), we aim to obtain a connected graph \( G = (V, E) \) to represent the structure of the story, where:
\begin{itemize}
    \item \( V \) is the set of vertices, each vertex representing a major event in the story.
    \item \( E \) is the set of edges, where each edge \( (u, v) \in E \) represents the temporal or causal relationship from event \( u \) to event \( v \).
\end{itemize}

For the definition of Edges E, We say Event A causes Event B if:
\begin{itemize}
    \item (the multi-factorial definition): in combination with other factors, Event A is a necessary or a sufficient condition for Event B \cite{Oppenheimer_Susser_2007}
    \item  (the probabilistic definition): the occurrence of Event A raises the probability of Event B occurring \cite{reichenbach1991direction}.
\end{itemize}

\section{Methodology}

Our complete Causal graph Model is an End-to-End model. We hope to input any story and generate a Connected Graph \(G\). This model contains four main parts:(1) Vertices Extraction, (2)Expert Index Extraction, (3) STAC Categorization, (4)Graph Construction.

\subsection{Vertices Extraction}

We define each vertex in our causal graph as a single event or state, represented by:
\[
V = \{v_1, v_2, \dots, v_n \mid v_i = \text{a \,single \,event/state}\}.
\]
These vertices serve as Vertices capturing key information with causal relationships in the narrative. Our goal is to transform the original text into concise, event-specific sentences by leveraging a LLM and prompt engineering. In particular, we used the LangChain framework to guide the LLM in generating simple sentences that reflect core plot elements. 

\paragraph{Requirements for Each Vertex}
\begin{enumerate}
    \item \textbf{Concise}: Each sentence must contain no more than two clauses.
    \item \textbf{Agent-Centered}: The subject (or agent) of the action must be explicitly identified, with only one subject per sentence.
    \item \textbf{Active Voice}: Each sentence should clearly convey an action initiated by its subject.
\end{enumerate}

\paragraph{Extraction Procedure}
We applied a structured prompting workflow to simplify the text into short, self-contained sentences, each representing a single narrative event:

\begin{enumerate}
    \item \textbf{Summarization}: The LLM receives a paragraph and generates a brief summary, ensuring each resulting sentence is as simple as possible.
    \item \textbf{Pronoun Substitution}: All pronouns are replaced with explicit referents. For a first-person narrative, the speaker is replaced by a clear identifier, such as the speaker's name or “The Protagonist” if none is provided.
    \item \textbf{Clause Simplification}: Complex or compound sentences are split into multiple simple sentences, each containing one core action or state. Unimportant details that do not affect the plot are removed.
    \item \textbf{Continuous Flow}: The resulting sentences are checked to ensure they preserve a logical, causal flow of events, discarding irrelevant or tangential information.
\end{enumerate}

By enforcing these requirements and following this workflow, we derive a set of concise, agent-specific sentences—each of which becomes a vertex in our causal graph. This method preserves the essential narrative structure while ensuring that each vertex encapsulates only a single event or state.

\subsection{Expert Index Extraction}

This section describes our methodology for extracting the \textit{Expert Index} features from each sentence and subsequently training a model to classify them. The first three features---Genericity, Eventivity, and Boundedness---adapt the genericity/eventivity/boundedness-habituality framework described by \citet{hemmatian-etal-2021-novel} and \citet{hemmatian-borujeni-2022-taking}; the remaining four features are included for the narrative causal-graph setting. We adopt seven key features grounded in traditional and computational linguistics literature; the full descriptions are provided in Table~\ref{Expert Index}:

\begin{enumerate} \item \textbf{Genericity}: Determines whether the sentence's subject is \textit{specific} (e.g., a person, a dog) or \textit{generic} (e.g., a season, an emotion) \cite{becker-etal-2017-classifying, Carlson1980}. \item \textbf{Eventivity}: Classifies the verb as \textit{dynamic} (observable actions such as speaking or running) or \textit{stative} (expressing states or non-action, such as deciding or thinking) \cite{becker-etal-2017-classifying, Vendler1967}. \item \textbf{Boundedness}: Identifies if a event is \textit{episodic} (occurs at a specific time), \textit{habitual} (recurring over time), or \textit{static} (always true or in a state of being) \cite{becker-etal-2017-classifying, Smith1991}. \item \textbf{Initiativity}: Distinguishes whether the subject \textit{initiates} the action (has agency) or \textit{receives} it (lacks agency) \cite{dai-huang-2018-building, Comrie1976}. \item \textbf{Time Start}: Notes if the event begins in the \textit{past} or the \textit{present} relative to the narrative timeline \cite{dowty1979word, Allen1983}. \item \textbf{Time End}: Determines if the event concludes in the \textit{present} or the \textit{future} \cite{dowty1979word, Allen1983}. \item \textbf{Impact}: Indicates whether the event’s effect persists (\textit{impact}) or is entirely \textit{resolved} by the time it ends \cite{dowty1979word, MoensSteedman1988}. \end{enumerate}

Except for \textbf{Boundedness}, which has three categories, each feature has two categories, for a total of 192 possible combinations. We refer to each resulting combination as an \textit{Expert Index}. Inspired by prior work that classified sentences as episodic, habitual, or static, we adopt a more granular approach to better capture distinctions relevant to our four main narrative labels: \textbf{Situation}, \textbf{Task}, \textbf{Action}, and \textbf{Consequence}.

To train a model for these features, we used RoBERTa, a robustly optimized variant of BERT\citep{liu2019roberta}. We prepared a dataset of 750 annotated sentences from 23 short stories and novel chapters, ensuring balanced coverage of tenses and narrative types. Human evaluations served as ground truth. The model was trained separately for each of the seven features and their respective categories, enabling transparent prediction of the \textit{Expert Index} for every sentence.

\subsection{STAC Categorization} 

We developed the STAC model to classify narrative sentences into four categories—\textbf{Situation}, \textbf{Task}, \textbf{Action}, and \textbf{Consequence}—based on structured thinking from business management. In practice, we observed that narrative events often follow a logical flow: a change in the environment (\textit{Situation}) prompts a requirement (\textit{Task}), leading to an activity (\textit{Action}), which in turn yields a lasting result (\textit{Consequence})\cite{minto_pyramid_2009}. Concretely:

\begin{enumerate} \item \textbf{Situation}: Provides background context or sets the stage for future events. \item \textbf{Task}: States an explicit requirement or responsibility that must be fulfilled. \item \textbf{Action}: Indicates an activity actively performed or just completed. \item \textbf{Consequence}: Describes the outcome of a prior event that changes the state. \end{enumerate}

To automatically assign these four STAC labels, we trained a model using both RoBERTa embeddings and \textit{Expert Index} features as inputs. Their relation is as follows: ~\ref{app: STAC Table}. Specifically, we extracted each sentence’s embedding from RoBERTa’s default Autotokenizer (a 768-length array), capturing semantic and contextual meanings. We then one-hot encoded the \textit{Expert Index} categories (non-ordinal attributes) to obtain binary vectors. By concatenating these embeddings and encoded features, we formed a comprehensive input array.

For classification, we used \textbf{XGBoost} due to its efficiency and robust performance relative to traditional models. The model was trained on human-labeled STAC categories and human-labeled \textit{Expert Index} features as ground truth, with regularization techniques to avoid overfitting. Once trained, the model can predict a sentence’s STAC category from its tokenized RoBERTa embedding and \textit{Expert Index} attributes.

\subsection {Graph Construction}
After classifying all vertices using the STAC model—\textbf{Situation}, \textbf{Task}, \textbf{Action}, and \textbf{Consequence}—we aimed to build a causal diagram capturing the complexity of narrative events. Initially, we considered 16 possible bonds (i.e., relationships) between the four STAC categories; however, only 11 of these bonds were meaningful in the actual narrative context. Furthermore, we observed that real-world events often exhibit relationships such as \textit{Action → Action} or \textit{Situation → Situation}, underscoring the non-linear nature of storytelling.

To systematically determine the edges between vertices, we adopted a five-iteration LangChain-based prompting process. This approach refines causal relationships in stages, ensuring that each edge is relevant, logically consistent, and supported by the narrative.

\paragraph{Iteration 1: STAC Bond Learning}
We first prompted the LLM to internalize the STAC bonding schema, which outlines valid causal connections among \textbf{Situation}, \textbf{Task}, \textbf{Action}, and \textbf{Consequence}. By learning these inherent relationships, the model could more accurately propose potential edges in subsequent steps.

\paragraph{Iteration 2: Causal Relation Identification}
Next, the LLM evaluated pairs of vertices (in total \(O(n^2/2)\) pairs) to propose potential causal links based on the STAC bonds. At this stage, the model only suggested edges that aligned with valid STAC relationships and logically connected one event’s outcome or state to another event’s occurrence.

\paragraph{Iteration 3: Logical Consistency and Pruning}
After generating an initial set of edges, the LLM applied counterfactual reasoning—asking, “If A did not occur, would B still happen?”—to filter out any bonds that did not have explicitly causal relationship. Non-causal or weakly supported edges were systematically pruned, leaving only robust causal connections.

\paragraph{Iteration 4: Isolated Vertices Refinement}
In the fourth step, the LLM revisited any vertices that remained isolated (i.e., lacking causal connections). By prompting the model with a “why” question, we explored whether there were overlooked causes or effects. If new connections surfaced, they were subjected to the same scrutiny and pruning as in Iterations 2 and 3, ensuring consistency and avoiding redundant links.

\paragraph{Iteration 5: Final Graph Construction}
Finally, the refined set of vertices and edges was compiled into a coherent graph that depicts the full range of causal relationships within the narrative. This final graph integrates all relevant Vertices and edges, with every link verified for logical soundness and alignment with the STAC bonding schema.

By iterating through these five steps, we resolved the complexities of linking narrative events—particularly cases where \textit{Action} leads to another \textit{Action} or \textit{Situation} follows another \textit{Situation}. The result is a structured causal diagram ~\ref{app:Example Graph} that accurately reflects the underlying relationships dictated by both the story and the STAC framework.

\section{Experiment Setup}\label{sec:experiment-setup}
\subsection{Corpus Collection}

We hand-collected excerpts from 50 full-length novels and 50 short stories, covering works published between \textit{1800 and 1950}. Each data selection features either one chapter from a novel or a complete short story, with lengths averaging 5,000 words. All narratives were sourced from various public domain web archives. These works were selected in part because our annotators were already familiar with the narratives, reducing ambiguity and enabling more consistent annotation.

The dataset incorporates both \textit{complete story cycles} (e.g., short stories) and \textit{fragmentary narratives} (e.g., chapters), allowing for comparative event-flow analysis \citep{sims2019literary, kirti2024enhancing}. Thematically, it spans \textit{fairy tales}, \textit{stream-of-consciousness storytelling} (e.g., Poe’s \textit{Berenice} \citep{poe1835berenice}), and \textit{implied-content stories} (e.g., works by O. Henry \citep{ohenry1906}), ensuring a diverse testing ground for event-extraction models \citep{levi2022detecting, elson2012dramabank}.

\subsection{Summarization and Dataset Structuring}

After selecting corpus material, we employ a multi-layered Large Language Model (LLM) pipeline to iteratively refine narrative content, forming our finalized corpus dataset. The pipeline extracts and refines key sentences and concepts based on the story's progression, creating a connected-event narrative structure. The input to the pipeline is a raw chapter or story from the gathered corpus material, and the output is a concise summarization where each sentence has a declarative, complete narrative structure \citep{goyal2022news, lu2023auto}.

After processing each piece in the corpus through the pipeline, we gather a dataset optimized for event flowchart mapping. The final summaries, averaging under 40 sentences for each short story or novel chapter, serve as standardized Vertices in the output graph. Details on the pipeline and prompt methodology are provided in the Appendix~\ref{app:Vertices Extraction}.

\subsection{Expert and STAC Labeling}\label{subsec:expert-stac-labeling}

To construct the event-flow graph, we apply a structured labeling process integrating \textit{expert index} classification and \textit{STAC labeling}. This ensures clear labeling of narrative components into actionable event Vertices \citep{barth2021annotation}.

We asked ten anonymous annotators to assign \textit{expert index} and \textit{STAC labeling} to every sentence in the dataset. When differences arose, the mode was used \citep{fleiss1971measuring}. Annotators assigned the \textit{expert index} based on predefined criteria introduced earlier. They were then instructed to assign \textit{STAC labeling} to the same sentences following a hierarchical rule set:
\begin{itemize}
    \item An execution of an action verb solely defines an \textit{action}.
    \item If no action verb is present, sentences implying an execution are labeled as \textit{tasks}.
    \item If a description is shaped by the main flow of events and tasks, it is a \textit{consequence}.
    \item Otherwise, it is classified as a \textit{situation}.
\end{itemize}
This layered process ensures consistency across the dataset, aligning narrative progression with structured event representation for final graph construction.

We also explored generating STAC labels and Expert Index levels using a standardized prompt driven by a Large Language Model (LLM), detailed explicitly in the Appendix~\ref{app:Vertices Extraction}. However, the resulting annotation performance was suboptimal. Specifically, after evaluation across 300 datasets compared to annotations produced by human annotators, the Cohen's Kappa \citep{cohen1960coefficient, landis1977measurement} for the Expert Index generated by the LLM was found to be 0.73, indicating good but not excellent agreement. In contrast, the Cohen's Kappa for STAC labels generated by the LLM fluctuated around 0.63, suggesting only moderate agreement and thus inadequate for reliable model training. Consequently, for all subsequent scenarios involving Expert Index and STAC labeling, we adopted human annotations exclusively as the ground truth.

\section{Experiments}

\subsection{Vertices Extraction Result}

We evaluate and compare the performance of different models by comparing and rating their performances on fifteen selected stories. Ten of these were short stories, and five were chapters from well-known novels: \textit{The Giver} \citep{lowry1993giver}, \textit{The Great Gatsby} \citep{fitzgerald1925greatgatsby}, and \textit{Rebecca} \citep{dumaurier1938rebecca}. For each story or chapter, three summaries were generated using the same prompt and parameter settings (detailed in the Appendix~\ref{app:STAC Categorization Unused},~\ref{app:Expert Index Unused}) with no post-editing, following standard practices for comparative evaluation of summarization models \citep{goyal2022news, lu2023auto}.

To reflect the downstream goal of transforming summaries into structured event flowcharts, we defined a three-part evaluation rubric based on existing summarization literature \citep{kryscinski2019neural, fabbri2021summeval}:
\begin{itemize}
    \item \textbf{Conciseness and Sentence Structure:} Clean sentence flow, minimal subordination, and avoidance of redundancy.
    \item \textbf{Coverage and Coherence:} Inclusion of all key story events in proper logical order.
    \item \textbf{Information Span \& Economy:} Avoidance of unnecessary elaboration or repeated ideas.
\end{itemize}

Each summary was scored across the three dimensions (0--5 scale per category, 15 max per summary) by three LLM models (GPT-4o, GPT-4 Turbo, Claude 3.5), and the mean was then taken. Two additional criteria, Agent-Centered and Active Voice, were achieved at 100\% by all models and thus not considered further in our analysis.

\begin{table}[!ht]
\centering
\begin{tabular}{|l|c|c|c|}
\hline
\textbf{Model} & \textbf{Concise} & \textbf{Cover} & \textbf{Info Span} \\
\hline
GPT-4o & 4.2 & 4.9 & 4.4 \\
GPT-4 Turbo & 3.9 & 4.7 & 4.5 \\
GPT-o1 & 4.1 & 4.4 & 4.2 \\
\hline
\end{tabular}
\caption{GPT-4o demonstrates superior performance across all evaluated dimensions.}
\label{tab:summary-eval-results}
\end{table}

These results suggest that GPT-4o consistently demonstrates superior performance, producing efficient narrative compression while retaining complete event arcs—a critical capability for generating effective, structured flowchart-ready summaries \citep{li2022title2event, sims2019literary}. Consequently, GPT-4o was selected as our primary summarization model for dataset structuring.

\subsection{Expert Index Result}
We used a RoBERTa-based classifier fine-tuned on a custom-labeled dataset of 1,000 summary-extracted sentences annotated by humans. The dataset was split 80/20 into training and testing sets, with hyperparameters tuned via default cross-validation. Each trait was modeled independently as a multi-class classification task.

Performance scores for each trait dimension are shown in Table~\ref{tab:trait-results}. Overall, the classifier exhibited strong performance on traits with more balanced or semantically distinct labels. \textit{Genericity}, \textit{Eventivity}, and \textit{Initiativity} all yielded F1-scores above 0.85 on their dominant classes. \textit{Boundedness} posed greater challenges due to conceptual overlap between the \textit{habitual} and \textit{static} classes, leading to reduced precision and recall.

The classifier achieved high overall accuracy across most traits, with particularly strong results for identifying Initiate vs. Receive references and dynamic event types. Errors in \textit{Boundedness} are unsurprising given the theoretical overlap between habitual and static categories. For traits with label imbalance, such as retextit{Time Start}, outcome reveals minor reduced recall.

\subsection{STAC Categorization Result}
We conducted a series of experiments on a dataset of 1,000 ground-truth annotated sentences to evaluate the effectiveness of incorporating Expert Index features for STAC classification. Each sentence in the dataset is labeled with one of four STAC categories (Situation, Task, Action, or Consequence). We used a standard train/test split (e.g., 80/20) and report the F1-score  for each category as well as the macro-averaged F1-score across all four labels. Six different classification models were compared to isolate the impact of the Expert Index (EI) features:

\begin{enumerate}\setlength{\itemsep}{0pt}
    \item \textbf{RoBERTa (sentence only)} – A baseline model using only RoBERTa sentence embeddings (768-dimensional) with a linear classifier.
    \item \textbf{RoBERTa + EI} – RoBERTa embeddings augmented with the 13-dimensional one-hot Expert Index vector (total 781 features) and classified by a linear layer.
    \item \textbf{XGBoost (EI only)} – An XGBoost classifier using only the 13 Expert Index features.
    \item \textbf{XGBoost (RoBERTa only)} – XGBoost using only 768-dim RoBERTa embedding as input.
    \item \textbf{XGBoost (RoBERTa + EI)} – XGBoost using the combined feature set of RoBERTa embedding + EI (781 features).
    \item \textbf{GPT-4 (prompt-based)} – Using GPT-4 directly for classification via prompt (zero-shot, without fine-tuning).
\end{enumerate}

\begin{figure}[t]
    \centering
    \includegraphics[width=\columnwidth]{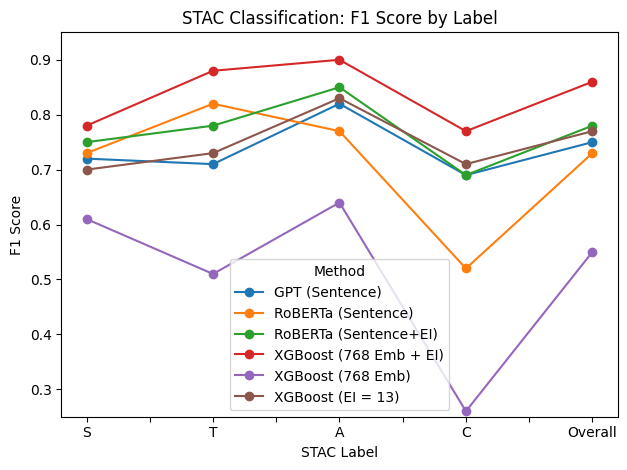}
    \caption{F1-score-score comparison across STAC labels for all six models. Each curve corresponds to a classification method, plotting F1-score for the four individual labels (S, T, A, C) and the overall macro-F1-score (rightmost point). The XGBoost model using both RoBERTa embeddings and Expert Index features (red curve) achieves the highest F1-score in every category.}
    \label{fig:stac_performance}
\end{figure}

As shown in Figure~\ref{fig:stac_performance}, models that incorporate the Expert Index features consistently outperform their counterparts that use only the sentence embedding. For instance, augmenting RoBERTa with the EI features raises the F1-score score in each category by at least 5 percentage points compared to using RoBERTa alone. This improvement is most pronounced for the \textit{Consequence} (C) category, where the RoBERTa+EI model achieves an F1-score of about 0.68 versus 0.55 with RoBERTa-only (a 13-point gain). Even the XGBoost classifier using only the 13 EI features (without any RoBERTa embedding) performs respectably across categories ($F_1\approx0.65$--$0.80$), underscoring that the Expert Index captures valuable signals for the STAC classification task.

Among all evaluated models, the XGBoost ensemble leveraging the combined RoBERTa + Expert Index features is the top performer. It attains the highest F1-score in each STAC category and the highest overall macro-F1-score. Notably, this model outperforms the GPT-4 classifier by approximately 10--15\% (relative) in F1-score score, and yields about a 30\% relative improvement over the baseline RoBERTa-only approach. These results demonstrate that incorporating the Expert Index not only consistently boosts classification accuracy for each STAC category, but that the combination of semantic embeddings with expert-driven features is especially powerful. The best model (XGBoost with RoBERTa+EI) provides a substantial performance margin over both a strong neural baseline and GPT-4, highlighting the benefit of hybridizing learned embeddings with expert knowledge.

\begin{table*}[t]
\centering
\begin{tabular}{lcc}
\toprule
\textbf{Dimension} & \textbf{Our Method vs GPT-4o} & \textbf{Our Method vs Claude 3.5} \\
\midrule
Causality vs. Chronology       & 100\% & 100\% \\
Explicit Motivations/Intent    & 95\%  & 92\%  \\
Granularity (Level of Detail)  & 86\%  & 84\%  \\
Logical Completeness           & 100\% & 100\% \\
Hierarchy or Grouping          & 94\%  & 92\%  \\
Accuracy of Connections        & 100\% & 100\% \\
Decision Points as Branches    & 97\%   & 95\%  \\
Ease of Reading                & 52\%  & 57\%  \\
\bottomrule
\end{tabular}
\caption{Win-rate of our model in pairwise comparisons against GPT-4o and Claude 3.5 on each dimension. Higher values indicate the percentage of cases where our model’s graphs were preferred for that dimension.}
\label{tab:dimcomparison}
\end{table*}

\subsection{Graph Formulation Result}
We define eight key dimensions for evaluating the quality of a causal event graph. Each dimension captures a different aspect of how well the graph represents the narrative’s causal structure:

\textbf{Causality vs. Chronology} – Does the graph emphasize true cause-effect relationships rather than merely the temporal order of events? Causal connectivity strongly shapes comprehension and recall of events \citep{Trabasso1985}.

\textbf{Explicit Motivations/Intent} – Are characters’ goals and intentions explicitly represented as causes for their actions? Agents’ motivations (the ``why’’ for actions) reflects the intentional dimension of narratives \citep{Zwaan1998} and ensures explanation on why events occur.

\textbf{Granularity (Level of Detail)} – Does the graph use an appropriate level of detail for events? A balanced level of detail enables both clarity and informativeness \citep{Mulkar2011}.

\textbf{Logical Completeness} – Are all necessary causal steps and connections present to form a logically complete story?  Missing links or unexplained leaps between events undermine narrative coherence \citep{Brewer1982}, undermining the logical soundness of the graph.

\textbf{Hierarchy or Grouping} – Does the graph organize events into higher-level groupings or hierarchical structures (e.g., subplots or phases)? A hierarchical organization (events grouped into episodes or goal-driven segments) improves understanding greatly \citep{Mandler1977}.

\textbf{Accuracy of Connections} – Are the causal links in the graph correct and faithful to the story? Each connection should reflect a true causal or enabling relation in the narrative, and incorrect causal links can mislead reasoning \citep{Pearl2009}. Every link in the graph shall not be coincidental nor erroneous.

\textbf{Decision Points as Branches} – Does the graph explicitly show branching at decision points? Representing decision points as branch Vertices highlights the narrative’s points of divergence (e.g., choices or hypothetical alternatives) and is important especially in interactive or non-linear narratives \citep{Moser2012}.

\textbf{Ease of Reading} – Is the graph easy to interpret visually, with a clear layout and labeling? Graph design principles (e.g., minimizing crossed links and clutter) improve human readability \citep{Purchase1997}, so a higher score means the graph is more reader-friendly.

\noindent\textbf{Experimental Setup.} We validated these evaluation dimensions by comparing our proposed method against strong baseline approaches, using large language models (LLMs) prompted to generate causal graphs from the same narratives. In particular, we benchmarked our method against GPT-4o and against Claude 3.5, as representative state-of-the-art LLMs~\ref{App: Eval_Prompt}. We also tested enhanced prompting with in-context examples: GPT-4o and Claude 3.5 denote prompting the LLM with 10 example narratives and their graphs (10-shot learning) to guide its generation. For each narrative text in our test set (100 narratives), both our method and a baseline LLM produced a causal graph. We then performed pairwise evaluations: for each narrative and each of the eight dimensions above, the graph from Method A was compared to the graph from Method B to decide which one was better along that specific dimension. This yields, per narrative, a binary win/loss outcome for each dimension. We conducted these pairwise comparisons for all relevant pairs: our method vs GPT-4o, our method and our method vs Claude 3.5, 

To ensure the reliability of the evaluation, we used a panel of five human annotators to judge the graph pairs dimension-by-dimension. Additionally, we employed an LLM-based evaluator (GPT-4) to perform the same pairwise judgments. We found a very high agreement between the aggregate human decisions and the LLM judge’s decisions: Cohen’s $\kappa = 0.92$ for dimension-level agreement. This suggests that the LLM-based evaluation is largely consistent with human, validating its use for scaling up our evaluation. In the analysis that follows, we thus report results based on the LLM evaluator’s judgments for all 100 narrative graph pairs, given the strong alignment with human annotators.

In Table~\ref{tab:dimcomparison}, we report the win-rates of our approach’s graphs compared to two baseline systems (GPT-4o and Claude 3.5) across the eight dimensions. The results show that our model substantially outperforms both baselines on almost all aspects of causal graph quality. Notably, it achieves near-100\% win rates against GPT-4o and Claude in dimensions such as \textit{Causality vs. Chronology}, \textit{Logical Completeness}, and \textit{Accuracy of Connections}, indicating that our graphs consistently capture causal structure, completeness, and correct links better than the baseline graphs. Similarly, high win-rate margins in \textit{Explicit Motivations}, \textit{Granularity}, and \textit{Hierarchy/Grouping} demonstrate the model’s strength in including character intents, appropriate detail, and structured organization of events. In contrast, for \textit{Ease of Reading}, the advantage of our model is much smaller (around 52–57\% win-rate), suggesting that the clarity and readability of our graphs are roughly on par with those generated by GPT-4o and Claude. Overall, these results highlight that our proposed graph formulation provides significant improvements in most qualitative dimensions of causal graph representation, while maintaining comparable readability.

\section{Conclusion} 
We have introduced a linguistics-focused, end-to-end approach for building causal graphs from narrative texts. By leveraging a lightweight \textit{Expert Index} to capture seven core linguistic traits, our STAC classifier improves both interpretability and accuracy in labeling events. A specialized, multi-step prompting strategy then constructs a logically consistent causal graph that outperforms GPT-4o and Claude 3.5 on most causal quality metrics. The results highlight the benefits of integrating interpretable feature engineering with modern language models for fine-grained causal reasoning. Our framework is open-source and readily adaptable for broader applications in summarization, discourse analysis, and knowledge graph construction.

\section*{Acknowledgments}

We thank Babak Hemmatian and Jared Hotaling for discussions and prior work on the genericity/eventivity/boundedness-habituality framework that informed the first three Expert Index dimensions.
We also thank all the anonymous reviewers for their constructive feedback and insightful comments.

\bibliography{custom}

\appendix

\section{Appendix}

\subsection{LLM Prompt for Vertices Extraction} \label{app:Vertices Extraction}
\begin{tcolorbox}[colback=gray!5!white, colframe=gray!80!black, sharp corners, boxrule=0.5pt, fontupper=\footnotesize\ttfamily, left=1mm, right=1mm, top=1mm, bottom=1mm]
\begin{enumerate}[leftmargin=*]
    \item I will input a paragraph to you and you need to do the following.
    \item You should summarize the sentences. All sentences should be SIMPLE sentences.
    \item If the story is told in first person POV, try to find out the speaker's name or something to refer to the speaker. If you really can't find anything, sub the speaker with 'The Protagonist'.
    \item Then, sub ALL pronouns, including the ones in the sentence, with the thing that they refer to.
    \item Then, Break ALL clauses into \textbf{SIMPLE SENTENCES}. Delete unimportant clause-level information. Be CONCISE.
    \item Your output at this time shall have LITTLE TO NO clauses.
    \item You need to check the sentences. If they contain clause, BREAK IT INTO TWO SENTENCES.
    \item The sentences, in their order, should give a continuous flow. DO NOT eliminate any important information that shows causal relationship.
    \item However, only information that pushes the plot/story is needed. Be concise and do not include ANY irrelevant information.
    \item Eventually, give me a summarization that focuses on causal relationships for the story.
\end{enumerate}
\end{tcolorbox}

\subsection{LLM Prompt for STAC Categorization (Unused)} \label{app:STAC Categorization Unused}
The following is our perspective on prompting as described in Section~\ref{sec:experiment-setup}, specifically in Subsection~\ref{subsec:expert-stac-labeling}. We attempted direct prompting using the STAC Model as we understood it; however, it did not serve as a suitable baseline. Instead, we employed it solely for comparison purposes.

\begin{tcolorbox}[colback=gray!5!white, colframe=gray!80!black, sharp corners=southwest, boxrule=0.5pt, left=1mm, right=1mm, top=1mm, bottom=1mm, fontupper=\footnotesize\ttfamily]
Classify each sentence in each chunk individually into either a situation, a task, an action or a consequence. Note that the sentences ARE NOT related.
We do these as follows:

1. Situation: Something that sets the stage of the BACKGROUND, without implying a particular action or task. The sentence will typically set the stage for something that happens later. Generally, it focuses on things that already happened at a certain stage of the story or something that would impact stuff later.

2. Task:  Describes an explicit requirement, want, or responsibility that needs to be fulfilled. The sentence would explicitly(the action’s name shall be mentioned)  mention some event that one subject would accomplish later, but hasn’t accomplished yet. If the sentence implies an action due to outforce changes, it’s categorized as a situation.

3. Action: This refers to an activity that is BEING or HAS JUST BEEN carried out by someone. It requires someone to ACTIVELY do the action. Otherwise, it shall be a situation or a consequence.

4. Consequence: Describes when something happens as a result of at least one thing prior AND has an everlasting impact. It’s always an action that ‘finishes’ (the action changed some state and does not normally change back) or a straightforward state change. It’s different from a situation by the fact that it should be a result of something mentioned before in the paragraph, whereas a situation happens spontaneously. 
\end{tcolorbox}

\subsection{LLM Prompt for Expert Index Extraction (Unused)} \label{app:Expert Index Unused}
The following is our perspective on prompting as described in Section~\ref{sec:experiment-setup}, specifically in Subsection~\ref{subsec:expert-stac-labeling}. We attempted direct prompting using the Expert Index Model as we understood it; however, it did not serve as a suitable baseline. So we used humans as the Baseline.

\begin{tcolorbox}[colback=gray!5!white, colframe=gray!80!black, sharp corners=southwest, boxrule=0.5pt, left=1mm, right=1mm, top=1mm, bottom=1mm, fontupper=\footnotesize\ttfamily]
IMPACT: 
I would give you a bunch of sentences and I want you to tell if the main event in the sentence has a lasting impact or if the main event is already resolved.
for instance:
- the door is left opened - impactful, focuses on shifting of door's state
-He opened the door. - resolved, focuses on the person
Border cases:
- If you cannot determine any main event from the sentence, mark it as resolved because of a lack of state of change. 
\end{tcolorbox}

\begin{tcolorbox}[colback=gray!5!white, colframe=gray!80!black, sharp corners=southwest, boxrule=0.5pt, left=1mm, right=1mm, top=1mm, bottom=1mm, fontupper=\footnotesize\ttfamily]
BOUNDEDNESS:
I would give you a bunch of sentences, not in any order,  and i want you to tell if the sentence's time span, labeled as 'Episodic', 'Habitual', or "Static'.

They are defined as follows:
- The event is Episodic if it happens only once And is at a specific time period (you may not know that period, but you know the period exists and has a bound)
- The Event is Habitual if the event happens on a regular basis. (There isn't a bound. The event is constant with intervals).
- The Event is Static if the Event describes a characteristic of the subject or if the event is constant and doesn't not have a clear bound. (Lacking Past OR future bound satisfies the category ). 
\end{tcolorbox}

\begin{tcolorbox}[colback=gray!5!white, colframe=gray!80!black, sharp corners=southwest, boxrule=0.5pt, left=1mm, right=1mm, top=1mm, bottom=1mm, fontupper=\footnotesize\ttfamily]
SPECIFICITY:
I would give you a bunch of sentences, not in any order,  and i want you to tell if the sentence has a proper noun or a common noun main subject, labeled as 'Specific' or 'Generic'. Define Strictly on the subject, not the implied subject. 

They are defined as follows:
- All proper nouns are Specific. We Treat 'The Protagonist' and Any type of PRONOUNS  as proper nouns in this case and are therefore Specific. Anything in First person POV is Specific. 
- Anything you can point to as 'It is THE ONE thing that does it' is Specific and treated as a proper noun. In a fairy tale, The Duck or A Tiger would be Specific because though they are not given a name, they act like proper nouns. (Think it like how the tiger's name would be Tiger)
- As an addition to 2, any live thing or personified thing the Starts with 'the' are treated as proper nouns and are thus Specific. 
- A common noun, when can STRICTLY trace back to proper noun
\end{tcolorbox}

\begin{tcolorbox}[colback=gray!5!white, colframe=gray!80!black, sharp corners=southwest, boxrule=0.5pt, left=1mm, right=1mm, top=1mm, bottom=1mm, fontupper=\footnotesize\ttfamily]
EVENTIVITY: 
I will give you a bunch of sentences. Classify each sentence in each chunk into either Stative, Dynamically Active or Mentally Active.
Do these as follows:
Check if the sentence describes a stative action (Labeled Stative). This includes possession(Have, consist, contain, etc.), thoughts(Think, remember, suspect, realize, etc.), senses(Feel, seem. etc.), and emotions that do not trigger an action (like, dislike, appreciate, etc.)

Or the sentence describes a dynamic action (Labeled Dynamically Active, which is characterized by more physical than mental movement). This includes the majority of the verbs(Jump, Walk, Suggest, Answer, etc.). Note that Talking or Expressing an opinion would be a dynamic action, because no mental action actually takes place. 

Or a mental action (Labeled Mentally Active). This includes action that happens mentally rather than physically, like decide, want, desire, hope, etc. 
\end{tcolorbox}

\begin{tcolorbox}[colback=gray!5!white, colframe=gray!80!black, sharp corners=southwest, boxrule=0.5pt, left=1mm, right=1mm, top=1mm, bottom=1mm, fontupper=\footnotesize\ttfamily]
TIME END: 
Classify each sentence in each chunk into either Time End Current (Label as C), Or Time End Future(Label as F).

We do these as follows:
Check if the Events will be continue happened after the sentence end itslef (In this case we label F(Future))
Def of End Future: 
A conclusion about what is happening now
(Things will continue [according to logic])
(Things will continue [for sure])
Things don't end with the statement.
\end{tcolorbox}

\begin{tcolorbox}[colback=gray!5!white, colframe=gray!80!black, sharp corners=southwest, boxrule=0.5pt, left=1mm, right=1mm, top=1mm, bottom=1mm, fontupper=\footnotesize\ttfamily]
TIME START:
Classify each sentence in each chunk into either Time Start Past, Or Time Start Now.

We do these as follows:
Check if the Events happened as we stated (In this case we label C(Current))
or the events happened as the sentences happened before (In this case we label P(Past))

If you find the event being persistent or stative and therefore does not have an explicitly start time, treat its start time as infinitely in the past and therefore label it as P. 
\end{tcolorbox}

\begin{tcolorbox}[colback=gray!5!white, colframe=gray!80!black, sharp corners=southwest, boxrule=0.5pt, left=1mm, right=1mm, top=1mm, bottom=1mm, fontupper=\footnotesize\ttfamily]
INITIATIVE:
I would give you a bunch of sentences, not in any order,  and i want you to tell if the sentence represents an action it initiates or Receives.
Define the main action and the main target through common sense and content. (NOT the subject). Now, I want you to tell me whether the target actively does(initiate), or receives an action(Receive). 
If the sentence itself is in passive form, it's automatically Receive.
If the sentence itself is in active form,  think about if the subject is able to do the action out of CHOICE or the action spontaneously happens. If the subject consciously does the action, it's an Initiate action. If not so, the subject Receives the action. 

app:STAC Categorization Unused
\end{tcolorbox}

\subsection{Table Description for the Expert Index} 

\begin{table*}[h!]
  \centering
  \begin{tabular}{| m{2cm} | m{2cm} | m{6cm} |}
  \hline
  \textbf{Features Name} & \textbf{Categories} & \textbf{Detail} \\ 
  \hline
  Generality & Specific & Refers to a particular instance or individual (e.g., a person, a dog). \\  
  \cline{2-3}
  & Generic & Refers to a general class or category (e.g., seasons, emotions).\\ 
  \hline
  Eventivity & Dynamic & Involves an observable action or change (e.g., speaking, running).\\ 
  \cline{2-3}
  & Stative & Describes a state of being or condition (e.g., deciding, thinking).\\ 
  \hline
  Boundness & Episodic & Refers to an event occurring at a specific time.\\ 
  \cline{2-3}
  & Habitual & Refers to actions that recur over time.\\ 
  \cline{2-3}
  & Static & Refers to something that is always true or a permanent state.\\ 
  \hline
  Time Start & Past & The event began in the past relative to the narrative moment.\\ 
  \cline{2-3}
  & Current & The event begins in the present relative to the narrative moment.\\ 
  \hline
  Time End & Current & The event concludes in the present relative to the narrative moment.\\ 
  \cline{2-3}
  & Future & The event will conclude in the future relative to the narrative moment.
\\ 
  \hline
  Initiality & Initiate & The subject has agency and initiates the action.\\ 
  \cline{2-3}
  & Receive & The subject passively receives the action, without agency.\\ 
  \hline
  Impact & Impactful & The event has a lasting or significant effect.\\ 
  \cline{2-3}
  & Resolved & The event's effect diminishes or resolves once completed.\\ 
  \hline
  \end{tabular}
  \caption{\label{Expert Index} Table Description for the Expert Index}
\end{table*}

\subsection{Example Graph of Our Method} \label{app:Example Graph}
\begin{figure*}[t!]
    \centering
    \includegraphics[width=0.7\textwidth,keepaspectratio]{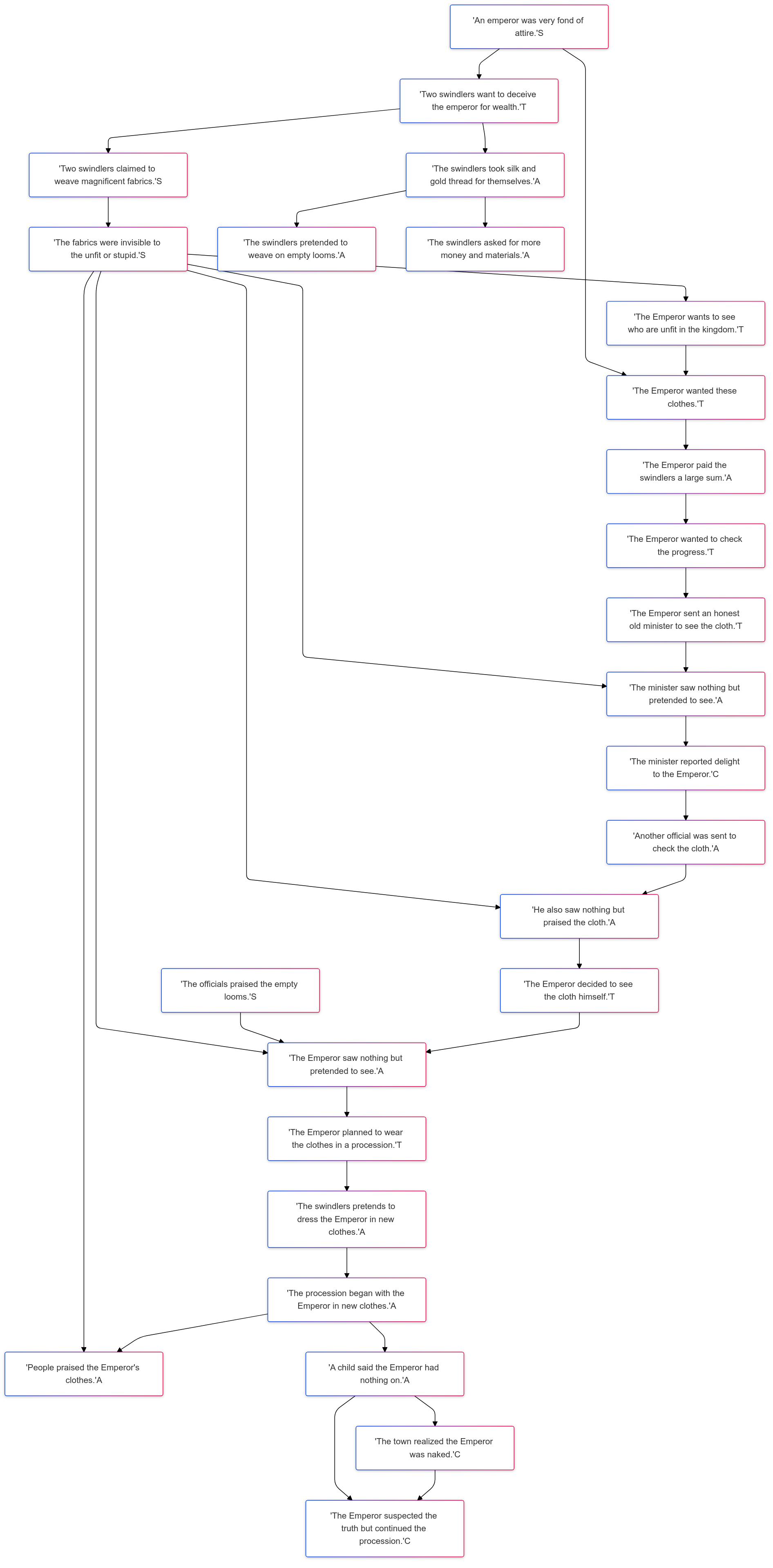}
    \caption{Example Graph Generation of Emperor's Cloth}
    \label{fig:demo_tables}
\end{figure*}

\subsection{Table Description for STAC Bonding} \label{app: STAC Table}

\begin{table*}[h!]
  \centering
  \begin{tabular}{| m{3cm} | m{3cm} | m{6cm} |}
  \hline
  \textbf{Begin Vertices} & \textbf{End Vertices} & \textbf{Definition} \\ 
  \hline
  Situation & Situation & The first situation may create a setting that directly influences or causes a change in another situation without any intermediate actions or tasks. \\  
  \cline{2-3}
  & Task & The current environment imposes certain responsibilities or actions on the agent. \\ 
  \cline{2-3}
  & Action & The environment itself drives the behavior, without an explicit task being identified first. \\ 
  \cline{2-3}
  & Consequence & The scenarios where background factors alone create significant changes in the state of affairs. \\   
  \hline
  Task & Action & This bond is a direct relationship where the execution of a task leads to a specific action.\\ 
  \cline{2-3}
  & Consequence & In this bond,  task itself will make an environment change as a result. \\ 
  \hline
  Action & Task/Action & This bond describes a sequence where one action leads directly to another action. Represents chains of immediate, active responses.\\ 
  \cline{2-3}
  & Consequence & This bond reflects a causal relationship where an act brings about a lasting change or outcome.\\ 
  \hline
  Consequence & Situation &  The consequence of a previous action or event sets up a new situation.(Different environment change)\\ 
  \cline{2-3}
  & Task/Action & The consequence directly drives the agent’s next move. \\ 
  \cline{2-3}
  & Consequence & This bond reflects a sequence of cascading outcomes, where one consequence leads to another. \\ 
  \hline
  \end{tabular}
  \caption{\label{STAC Bonding} Table Description for STAC Bonding}
\end{table*}

\subsection{Expert Index Result} \label{app: Expert Index Result}

\begin{table}[ht]
\centering
\begin{tabular}{|l|c|c|c|c|}
\hline
\textbf{Label} & \textbf{Precision} & \textbf{Recall} & \textbf{F1}\\
\hline
\textit{Genericity} (Generic) & 0.72 & 0.58 & 0.64\\
\textit{Genericity} (Specific) & 0.93 & 0.96 & 0.94\\
\hline
\textit{Eventivity} (D.Active) & 0.94 & 0.93 & 0.93\\
\textit{Eventivity} (M.Active) & 0.68 & 0.92 & 0.7\\
\textit{Eventivity} (Stative) & 0.85 & 0.75 & 0.80 \\
\hline
\textit{Boundedness} (Ep.) & 0.92 & 0.88 & 0.90\\
\textit{Boundedness} (Hab.) & 0.31 & 0.36 & 0.33\\
\textit{Boundedness} (Static) & 0.73 & 0.80 & 0.76\\
\hline
\textit{Initiativity} (Initiate) & 0.91 & 0.89 & 0.90\\
\textit{Initiativity} (Receive) & 0.84 & 0.86 & 0.85\\
\hline
\textit{Time End} (Present) & 0.92 & 0.86 & 0.89 \\
\textit{Time End} (Future) & 0.63 & 0.78 & 0.69 \\
\hline
\textit{Time Start} (Past) & 0.96 & 1.00 & 0.98 \\
\textit{Time Start} (Present) & 1.00 & 0.60 & 0.73\\
\hline
\textit{Impact} (Impactful) & 0.88 & 0.76 & 0.82 \\
\textit{Impact} (Resolved) & 0.84 & 0.89 & 0.87 \\
\hline
\end{tabular}
\caption{Classification results (test set, $n=200$) for each trait and class label.}
\label{tab:trait-results}
\end{table}

\label{sec:appendix}

\subsection{Evaluation of Causal Graph Prompt} \label{App: Eval_Prompt}
\begin{tcolorbox}[breakable, colback=gray!5!white, colframe=gray!80!black, sharp corners=southwest, boxrule=0.5pt, left=1mm, right=1mm, top=1mm, bottom=1mm, fontupper=\footnotesize\ttfamily]
Input Story: xxxxX
Causal Graph 1: xxxxxx
Causal Graph 2: xxxxxx

Your job is to make judgement for each of the Causal Graph, determine which one is better in each of the dimension, here is the dimension description:
\begin{enumerate}[leftmargin=*]
    \item Causality vs. Chronology: Does the diagram emphasize actual cause-and-effect rather than merely stringing events in time?
    \item Explicit Motivations/Intent: Are the driving reasons (e.g., revenge, pride, fear) clearly shown so the reader sees why a character or force triggers the next event?
    \item Accuracy of Connections: Do arrows represent genuine causal links (A enables or drives B), and are there any missing or spurious connections?
    \item Clarity and Brevity of Nodes: Are node labels concise and unambiguous? Too much text can clutter the diagram and obscure the causal flow.
    \item Granularity/Level of Detail:Is the diagram capturing just enough detail to show cause-effect without trivial or irrelevant steps?
    \item Logical Completeness: Does it include all critical causes and effects for key outcomes, so nothing pivotal is left out?
\end{enumerate}

\end{tcolorbox}

\end{document}